# Lifeworld Analysis

**Philip Agre**                                                                      PAGRE@UCSD.EDU
*Department of Communication 0503*
*University of California, San Diego*
*La Jolla, CA 92093, USA*

**Ian Horswill**                                                                     IAN@ILS.NWU.EDU
*Northwestern University Computer Science Department*
*1890 Maple Avenue*
*Evanston, IL 60201, USA*

## Abstract

We argue that the analysis of agent/environment interactions should be extended to include the conventions and invariants maintained by agents throughout their activity. We refer to this thicker notion of environment as a *lifeworld* and present a partial set of formal tools for describing structures of lifeworlds and the ways in which they computationally simplify activity. As one specific example, we apply the tools to the analysis of the TOAST system and show how versions of the system with very different control structures in fact implement a common control structure together with different conventions for encoding task state in the positions or states of objects in the environment.

## 1. Introduction

Biologists have long sought concepts to describe the ways in which organisms are adapted to their environments. Social scientists have likewise sought concepts to describe the ways in which people become acculturated participants in the social worlds around them. Yet it has been difficult to approach these phenomena with the methods of computational modeling. We can see at least two reasons for this difficulty. The first is that the tradition of modeling in artificial intelligence developed around a concern with cognition, that is, mental processes understood to intervene between stimuli and responses in human beings. Although minority traditions such as ecological psychology reacted against this approach to studying human life, they have not been able to translate their concepts into computational mechanisms that match the expressive power of symbolic programming. The second reason is more subtle: if one conceives both organisms and their environments as spatially extended mechanisms that can be explained according to the same principles then the boundary between them (the surface of the body) is not particularly different from, or more interesting than, the rest of the total organism-environment system. The challenge for computational modeling, then, is to conceptualize agents' adaptations to their environments in ways that neither treat agents as isolated black boxes or dissolve them into one big machine.

For these purposes, we find it useful to distinguish between two aspects of an agent's involvement in its familiar environment: its embodiment and its embedding. "Embodiment" pertains to an agent's life as a body: the finiteness of its resources, its limited perspective on the world, the indexicality of its perceptions, its physical locality, its motility, and so on. "Embedding" pertains to the agent's structural relationship to its world: its habitual paths, its customary practices and how they fit in with the shapes and workings of things, its connections to other agents, its position in a set of roles or a hierarchy, and so forth. The concept of embedding, then, extends from more concrete kinds of locatedness in the world (places, things, actions) to more abstract kinds of location (within social systems, ecosystems, cultures, and so on). Embodiment and embedding are obviously interrelated, and they each have powerful consequences both for agents' direct dealings with other agents and for their solitary activities in the physical world. Our principal focus in this article is on embedding, and particularly on the ways in which agents maintain relationships to objects that are functionally significant for their tasks.

In this paper we develop some thoughts about embodiment and embedding as follows:

- Section 2 reviews the concept of the environment as it developed with the early work of Newell and Simon.
- Section 3 introduces our own adaptation of the traditional idea, which we call *lifeworlds*, and we sketch what is involved in *lifeworld analysis*.
- Section 4 introduces informally the concept of *factorization* of lifeworlds; this refers roughly to the structures of the lifeworld that permit agents' decisions to be made independently of one another.
- Section 5 defines the basics of our formal theory of lifeworld analysis, namely the concepts of environments, actions, policies, factorization, and the reduction of one environment to another. The purpose of this formalism is to characterize the kinds of interactions that can arise between agents and their familiar lifeworlds.
- Section 6 briefly introduces the computer program we wrote to illustrate some of the phenomena of lifeworlds.
- Section 7 then applies our formalism to modeling the world in which our program operates; it proceeds by modeling successively more complicated versions of this world.
- Section 8 explains how our program keeps track of the objects in the world that figure in its activities, and discusses the issues that arise when trying to model this keeping-track in formal terms.
- Section 9 sums up our formal work by explaining the precise relationship between the program and the formal model of its world.
- Section 10 then expands our theory of lifeworlds more informally by introducing the concept of *cognitive autopoiesis*, which is the collection of means by which agents manipulate their surroundings to provide the conditions of their own cognitive processes; we provide a taxonomy of these phenomena.
- Section 11 concludes by suggesting some directions for future work.

## 2. The Concept of the Environment

Intuitively, the notion of "the environment" in AI and robotics refers to the relatively enduring and stable set of circumstances that surround some given individual. My environment is probably not the same as yours, though they may be similar. On the other hand, although my environment starts where I leave off (at my skin, perhaps), it has no clear ending-point. Nor is it necessarily defined in terms of metric space; if physically distant circumstances have consequences for my life (via the telephone, say) then they are properly regarded as part of my environment as well. The environment is where agents live, and it determines the effects of their actions. The environment is thus a matter of importance in computational modeling; only if we know what an agent's environment is like can we determine if a given pattern of behavior is adaptive. In particular we need a *positive theory* of the environment, that is, some kind of principled characterization of those structures or dynamics or other attributes of the environment *in virtue of which* adaptive behavior is adaptive.

Herbert Simon discussed the issue in his pre-AI work. His book *Administrative Behavior* (1947), for example, presents the influential theory that later became known as *limited rationality*. In contrast to the assumption of rational choice in classical economics, Simon describes a range of cognitive limitations that make fully rational decision-making in organizations impracticable. Yet organizations thrive anyway, he argues, because they provide each individual with a structured environment that ensures that their decisions are good enough. The division of labor, for example, compensates for the individual's limited ability to master a range of tasks. Structured flows of information, likewise, compensate for the individual's limited ability to seek this information out and judge its relevance. Hierarchy compensates for the individual's limited capacity to choose goals. And fixed procedures compensate for individuals' limited capacity to construct procedures for themselves.

In comparison to Simon's early theory in *Administrative Behavior*, AI has downplayed the distinction between agent and environment. In Newell and Simon's early work on problem solving (1963), the environment is reduced to the discrete series of choices that it presents in the course of solving a given problem. The phrase "task environment" came to refer to the formal structure of the search space of choices and outcomes. This is clearly a good way of modeling tasks such as logical theorem-proving and chess, in which the objects being manipulated are purely formal. For tasks that involve activities in the physical world, however, the picture is more complex. In such cases, the problem solving model analyzes the world in a distinctive way. Their theory does not treat the world and the agent as separate constructs. Instead, the world shows up, so to speak, phenomenologically: in terms of the differences that make a difference for *this* agent, given its particular representations, actions, and goals. Agents with different perceptual capabilities and action repertoires, for example, will inhabit different task environments, even though their physical surroundings and goals might be identical.

Newell and Simon's theory of the task environment, then, tends to blur the difference between agent and environment. As a framework for analysis, we find the phenomenological approach valuable, and we wish to adapt it to our own purposes. Unfortunately, Newell and Simon carry this blurring into their theory of cognitive architecture. They are often unclear whether problem solving is an activity that takes place wholly within the mind, or whether it unfolds through the agent's potentially complicated interactions with the physical world. This distinction does not arise in cases such as theorem-proving and chess, or in any other domain whose workings are easily simulated through mental reasoning. But it is crucial in any domain whose actions have uncertain outcomes. Even though we wish to retain Newell and Simon's phenomenological approach to task analysis, therefore, we do not wish to presuppose that our agents reason by conducting searches in problem spaces. Instead, we wish to develop an analytical framework that can guide the design of a wide range of agent architectures. In particular, we want an analytical framework that will help us design the simplest possible architecture for any given task.

## 3. Lifeworlds

We will use the term *lifeworld* to mean an environment described in terms of the customary ways of structuring the activities that take place within it — the conventional uses of tools and materials, the "loop invariants" that are maintained within it by conventional activities, and so on. The term originally comes from phenomenological sociology (Schutz & Luckmann, 1973), where it refers to the familiar world of everyday life, and specifically to that world as described in the terms that make a difference for a given way of life. Cats and people, for example, can be understood as inhabiting the same physical environment but different lifeworlds. Kitchen cupboards, window sills, and the spaces underneath chairs have different significances for cats and people, as do balls of yarn, upholstery, television sets, and other cats. Similarly, a kitchen affords a different kind of lifeworld to a chef than to a mechanic, though clearly these two lifeworlds may overlap in some ways as well. A lifeworld, then, is not just a physical environment, but the patterned ways in which a physical environment is functionally meaningful within some activity.

This idea is similar to Gibson's theory of perception (1986), but the two theories also differ in important ways. Whereas Gibson believes that the perception of worldly affordances is direct, we believe that the perceptual process can be explained in causal terms. Also, whereas Gibson treated the categories of perception as essentially biological and innate, we regard them as cultural and emergent.

In analyzing a lifeworld, one attempts to draw out the individual structures within it that facilitate its customary activities. For example, lifeworlds typically contain artifacts such as tools that have been specifically evolved to support those activities. These tools are also arranged in the world in ways that simplify life and reduce the cognitive burden on individuals: cups are typically found in cupboards, food in refrigerators and grocery stores. No one needs to remember where butter is found in a specific grocery store because butter in all grocery stores is found in a well-defined dairy section, usually along a wall, which can be recognized from a distance; once the dairy section is in view, the butter will be visible in a definite area. Artifacts are also designed to make their functional properties perceptually obvious. Handles are perceptibly suited for picking up, knobs are perceptibly suited for turning, forks are perceptibly suited for impaling things, and so on (Brady, Agre, Braunegg, & Connell, 1984; Winston, Binford, Katz, & Lowry, 1983). Contrarily, *it can generally be assumed that artifacts that provide no readily perceptible grounds for drawing functional distinctions are in fact interchangeable.* Usually, when some functionally significant property of an object is not obvious, the lifeworld provides some alternate way of marking it. If you see a record player in my house, for example, then you will assume that it is mine unless you have some specific reason not to. These aspects of lifeworlds tend to make it easy to perform particular kinds of activities within them without having to remember too many facts or reinvent the screwdriver from first principles.

Lifeworlds contain networks of interacting conventions and practices that simplify specific aspects of specific activities. The practices relieve agents of the burden of solving certain problems on the spot and diffuse their solutions throughout the activity of the agent or of many agents. For example, a hospital might try to get along without maintaining sterile conditions. People always have germs, so technically they are always infected. The problem is making sure that those infections never get out of control. The most direct solution would be constantly to monitor patients, assess their degree of infection and treat them when it becomes severe. Since this is undesirable for any number of reasons, a hospital instead tries to *prevent* infections in patients by maintaining sterile conditions. They might do this, for example, by looking for contaminated objects and surfaces and disinfecting them. Unfortunately, sterility is not a visible surface characteristic. Instead, hospitals solve the problem by *structuring space and activity*. Different locations are kept more or less sterile depending on their conventional uses: operating rooms are more sterile than hallway floors. Objects that can generate germs (people) are washed, masked, and gloved. Critical instruments that come in contact with them are specially sterilized after use. Tongue depressors are assumed to be dirty when they are in the trash can (or biohazard bag) and clean when they are wrapped in paper. All objects and surfaces are periodically disinfected regardless of their level of contamination. These practices are maintained regardless of the immediate need for them. If a hospital were (for some reason) to temporarily find itself without any patients, its workers would not stop washing their hands or disinfecting the bathrooms.

## 4. Factorization of Lifeworlds

Simon, in *Sciences of the Artificial* (1970), argued that complex systems had to be "nearly decomposable." His model for this was the rooms in a building, whose walls tend to minimize the effects that activity in one room has upon activity in another. Sussman (1975), in his analysis of block-stacking tasks, classified several types of "subgoal interactions" that result from attempts to break tasks down into subtasks; one hopes that these tasks will be decomposable, but bugs arise when they are not decomposable enough. One assumes that a task is decomposable unless one has reason to believe otherwise. Sussman's research, and the rich tradition of planning research that it helped inaugurate, concerned the difficult problem of constructing plans in the presence of subgoal interactions. Our goal, complementary to theirs, is to analyze the many ways in which tasks really are decomposable, and to derive the broadest range of conditions under which moment-to-moment activity can proceed without extensive analysis of potential interactions.

A non-pathological lifeworld will be structured in ways that limit or prevent interactions among subtasks. Some of these structures might be taxonomized as follows:

- *Activity partition.* Most lifeworlds separate activities under discrete headings: sewing is a distinct activity from bathing, gathering food is a separate activity from giving birth, and so on. These distinctions provide the basis for reckoning "different activities" for the purposes of most of the rest of the partitions. The boundaries among the various activities are often marked through some type of ritual.

- *Spatial partition.* Different things are often done in different places. Tasks may be confined to the places where their associated tools or materials are stored, or where suitable conditions or lighting or safety obtain. These places may even be close together, as when different recipes are prepared in different sections of countertop space or different kinds of food are kept in different parts of one's plate, with boundary regions perhaps employed to assemble forkfuls of neighboring foods. In general, activities are arranged in space, and decisions made in one place tend to have minimal interaction with decisions made in other places. Of course spatial distance brings no absolute guarantees about functional independence (using up all the resources at one location will prevent them from being carted to another location for another use later on), so these are just general tendencies.

- *Material partition.* Different activities often involve different materials, so that decisions that affect the materials in one activity do not interact with decisions that affect the materials of the other activity.

- *Temporal partition.* Different activities often take place at different times, thus limiting the channels through which they might constrain one another. These times might be standardized points during the cycle of the day or week, or their ordering might be constrained by some kind of precondition that the first activity produces and successive ones depend upon.

- *Role partition.* Simon pointed out that division of labor eases cognitive burdens. It does this in part by supplying individuals with separate spheres in which to conduct their respective activities.

- *Background maintenance.* Many activities have background conditions that are maintained without reference to specific goals. For example, one maintains stocks of supplies in the pantry, puts things back where they belong, and so forth. Hammond, Converse, and Grass (1995) call some of these "stabilization." (See Section 5.) What these practices stabilize are the relationships between an agent and the materials used in its customary activities. They tend to ensure, for example, that one will encounter one's hammer or the currently opened box of corn flakes in definite sorts of recurring situations. They thus reduce the complexity of life, and the variety of different hassles that arise, by encouraging the rise of routine patterns and cycles of activity rather than a constant stream of unique puzzles.

- *Attributes of tools.* Numerous properties of tools limit the interactions among separate decisions. Virtually all tools are resettable, meaning that regardless of what one has been doing with them, they can be restored to some normal state within which their full range of functionalities is accessible. (This of course assumes that one has only been using the tools in the customary ways and has not been breaking them.) Thus the properties of the tool do not place any ordering constraints on the activities that use it. Likewise, most tools are not committed to tasks over long periods. Once you have turned a screw with a screwdriver, for example, the screwdriver does not stay "stuck" to that screw for any long period. Thus it is not necessary to schedule the use of a screwdriver unless several people wish to use it at once. Exceptions to this general rule include bowls (whose ingredients must often sit waiting for future actions or conditions, and which cannot contain anything else in the meantime), stove burners (which sometimes must remain committed to heating particular dishes until they have reached certain states and not before), and clamps (which must remain fastened until the glue has dried or the sawing operations have been completed).

- *Supplies of tools.* These latter tools raise the spectre of generalized scheduling problems and the potential for deadlock among multiple activities, and such problems do in fact sometimes arise when cooking for more people than the number to which a given kitchen is adapted. Most of the time, though, one solves such problems not through scheduling but simply through having enough of the tools that must remain committed to particular purposes over a period of time. Lansky and Fogelsong (1987) modeled the effects on search spaces of limited interactions between different cooks using overlapping sets of tools.

- *Warning signs.* When things go wrong, unpleasant subgoal interactions can ensue. To avoid such difficulties, an individual, community, or species keeps track of warning signs and cultivates the capacity to notice them; these warning signs include supplies running low and funny smells. This is often done on a primitive associative level, as when rats stay away from smells that were associated with stuff that made them sick or people develop phobias about things that were present when they suffered traumas. Communities often arrange for certain warning signs to become obtrusive, as when kettles whistle or natural gas is mixed with another gas that has a distinctive smell.

- *Simple impossibility.* Sometimes things are just impossible, and obviously so, so that it is not necessary to invest great effort in deciding not to do them.

- *Monotonicity.* Many actions or changes of state are irreversible. Irreversible changes cause decisions to interact when certain things must be done before the change takes place. But it also provides a structure for the decision process: the lifeworld needs to make it evident *what* must be done before a given irreversible change occurs.

- *Flow paths.* Often a lifeworld will be arranged so that particular materials (parts on an assembly line, paperwork in an organization, food on its way from refrigerator to stove to table) follow definite paths. These paths then provide a great deal of structure for decision-making. By inspecting various points along a path, for example, one can see what needs to be done next. Or by determining where an object is, one can determine what must be done to it and where it must be taken afterward. Some of these paths are consciously mapped out and others are emergent properties of a set of customs.

- *Cycles.* Likewise, many lifeworlds involve stable cycles of activities, perhaps with some of the cycles nested inside of others. The resulting rhythms are often expressed in recurring combinations of materials, decisions, spatial arrangements, warning signs, and so on.

- *Externalized state.* To computer people, "state" (used as a mass noun) means discernible differences in things that can be modified voluntarily, and that can be interpreted as functionally significant in some way. Early AI did not treat internal state (memory) and external state (functionally significant mutable states of the world) as importantly different, and it is often analytically convenient to treat them in a uniform fashion. It is often advantageous to record state in the world, whether in the relative locations of things and the persistent states (in the count noun sense) that they are left in (Beach, 1988). For example, one need not remember whether the eggs have been broken if that fact is readily perceptible, if one's attention will be drawn to it on a suitable occasion, and if one understands its significance for the task. Likewise, one can save a great deal of memory by retrieving all of the ingredients for an evening's recipes from the cupboards and placing them in a customary place on the shelf.

Lifeworlds, then, have a great deal of structure that permits decisions to be made independently of one another. The point is not that real lifeworlds permit anyone to live in a 100% "reactive" mode, without performing any significant computation, or even that this would be desirable. The point, rather, is that the nontrivial cognition that people do perform takes place against a very considerable background of familiar and generally reliable dynamic structure.

The factorability of lifeworlds helps particularly in understanding the activities of an agent with a body. A great deal of focusing is inherent in embodiment. If you can only look in one place at a time, or handle only one tool at a time, your activities will necessarily be serial. Your attention will have a certain degree of hysteresis: having gotten to work on one countertop or using one particular tool, for example, the most natural step is to carry on with that same task. It is crucial, therefore, that different tasks be relatively separate in their consequences, that the lifeworld provide clues when a change of task is necessary, and that other functionally significant conditions can generally be detected using general-purpose forms of vigilance such as occasionally looking around. Of course, certain kinds of activities are more complex than this, and they require special-purpose strategies that go beyond simple heuristic policies such as "find something that needs doing and do it." The point is that these more complex activities with many interacting components are rare, that they are generally conducted in specially designed or adapted lifeworlds, and that most lifeworlds are structured to minimize the difficulty of tasks rather than to increase it.

These various phenomena together formed the motivation for the concept of indexical-functional or deictic representation (Agre & Chapman, 1987; Agre, 1997). Embodied agents are focused on one activity and one set of objects at a time; many of these objects are specifically adapted for that activity; their relevant states are generally readily perceptible; objects which are not perceptibly different are generally interchangeable; and stabilization practices help ensure that these objects are encountered in standardized ways. It thus makes sense, for most purposes, to represent objects in generic ways through one's relationships to them. The flashlight I keep in the car is *the-flashlight-I-keep-in-the-car* and not FLASHLIGHT-13. I maintain a stable relationship to this flashlight by keeping it in a standard place, putting it back there when I am done with it, using it only for its intended purposes, keeping its batteries fresh, and so on. Its presence in the environment ensures that I have ready access to light when my car breaks down at night, and therefore that I need not separately plan for that contingency each time I drive. The conventional structures of my own activity maintain the flashlight's presence as a "loop invariant." Both the presence of the flashlight and the activities that ensure it are structures of my lifeworld.

## 5. Environments, Policies, and Reducibility

In this section, we will introduce our formalism. The purpose of the formalism is not directly to specify the workings of the agent's cognitive machinery. Instead, its purpose is to construct "principled characterizations of interactions between agents and their environments to guide explanation and design" (Agre, 1995). The formalism, in other words, describes an agent's embodied activities in a particular environment. Having characterized the dy-namics of those activities, it becomes possible to design suitable machinery. As a matter of principle, we want to design the simplest possible machinery that is consistent with a given pattern of interaction (Horswill, 1995). We therefore make no *a priori* commitments about machinery. We do not favor any particular architecture until a particular activity has been analyzed. Nor do we make any *a priori* commitments about matters such as analog versus digital, "planning" versus "reaction," and so on. Our experience has been that real lifeworlds and real activities incorporate a great deal of useful dynamic structure, and that any effort we invest in studying that structure will be repaid in parsimonious theories about machinery. But we intend our methods to be equally useful for investigating all types of activity and designing all types of machinery that might be able to participate in them.

The concept of a lifeworld will not appear as a specific mathematical entity in our formalism. The intuition, however, is this: while there is an objective material environment, the agent does not directly deal with all of this environment's complexity. Instead it deals with a *functional* environment that is projected from the material environment. That projection is possible because of various conventions and invariants that are stably present in the environment or actively maintained by the agent. The lifeworld should be understood as this functional world together with the projection and the conventions that create it. This section summarizes the formal model of environmental specialization given by Horswill (1995); for proofs of the theorems, see the original paper. Subsequent sections will apply and extend the model.

We will model environments as state machines and the behavior of agents as policies mapping states to actions.

- An *environment* $E$ is a pair $(S, A)$ where $S$ is its state-space and $A$ its set of possible actions.

- An *action* $a: S \to S$ is a mapping from states to states.

- A *policy* $p: S \to A$ is a mapping from states to actions to be taken. In this paper, the states will only include facts about the physical environment, but it is a straightforward matter to include an agent's internal states as well (Horswill, 1995).

The combination of a policy with an environment creates a dynamic system: the environment's state is mapped by the policy to an action that maps the environment to a new state and the whole process is repeated.

- A *discrete control problem* (DCP) is a pair $(E, G)$ of an environment $E$ and a goal $G$, which is some subset of $E$'s state space.

- A policy *solves* the problem if the dynamic system it generates with the environment eventually reaches a goal state.

- It solves the problem *and halts* if it remains within $G$ once entering it.

For example, consider a robot moving along a corridor with $n$ equally spaced offices labeled 1, 2, 3, and so on. We can formalize this as the environment $\mathcal{Z}_n = (\{0, 1, ..., n-1\}, \{inc_n, dec, i\})$, where $i$ is the identity function, and where $inc_n$ and $dec$ map an integer $i$ to $i+1$ and $i-1$, respectively, with the proviso that $dec(0) = 0$ and $inc_n(n-1) = n-1$ (see Figure 1). Note that the effect of performing the identity action is to stay in the same state.

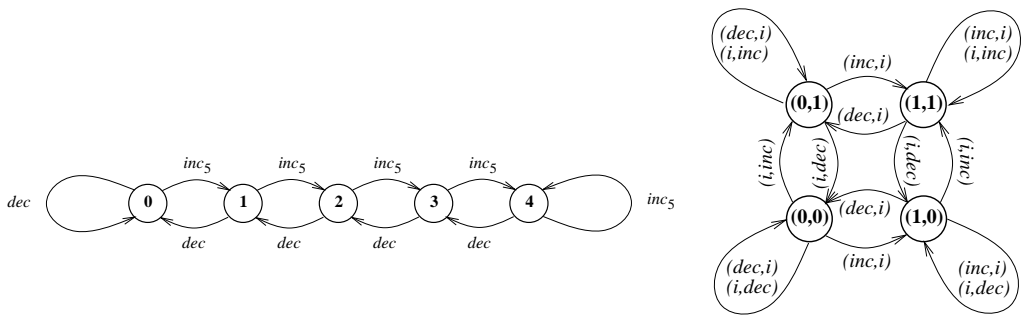

Figure 1: The environment $Z_5$ (left) and and the serial product of $\mathcal{Z}_2$ with itself, expressed as graphs. Function products have been written as pairs, *i.e.* $inc{\times}i$ is written as $(inc, i)$. Identity actions ($i$ and $i{\times}i$) have been left undrawn to reduce clutter.

We emphasize that *a policy is a model of an agent's behavior, not of the causal/computational processes by which that behavior is exhibited*. It specifies *what* an agent does in each state, not *how* it does it. It is thus a theoretical construct, not a data structure or algorithm in the agent's head. We will examine the implementation issues that surround policies in section 8.

### 5.1 Product Environments

The majority of the formal sections of this paper will explore the phenomenon of factoring. In particular, we will explore how policies for factorable environments can be composed from policies for the factors. In state-machine models of environments, factorization is the factorization of the state-space; the environment's state-space is a Cartesian product of other state-spaces. The environment, as a whole, is "factorable" into its component sub-environments. For example, the position of the king on a chess board has row and column components. It can be thought of as the "product" of those components, each if which is isomorphic to $\mathcal{Z}_8$ (since there are eight rows and eight columns). If we consider an environment in which a car drives through an 8×8 grid of city blocks, we see that it too is a kind of product of $\mathcal{Z}_8$ with itself. Both environments have 8×8 grids as state spaces, but the car environment only allows one component to change at a time, whereas the king environment allows both to change.

We must therefore distinguish different *kinds* of factorization. We will call the chessboard case the *parallel product* of $\mathcal{Z}_8$ with itself, while the car case is its *serial product*. We will focus on another kind of factorization later. Let the Cartesian product of two functions $f$ and $g$ be $f{\times}g\colon (a,\ b) \mapsto (f(a),\ g(b))$, and let $i$ be the identity function. For two environments $E_1 = (S_1, A_1)$ and $E_2 = (S_2, A_2)$, we will define the parallel product to be

$$E_1 \,\|\, E_2 \;=\; (S_1{\times}S_2, \{a_1{\times}a_2 : a_1 \in A_1, a_2 \in A_2\})$$

LIFEWORLD ANALYSIS

and the serial product to be

$$E_1 \rightleftharpoons E_2 \;=\; (S_1 \times S_2, \{a_1 \times i : a_1 \in A_1\} \cup \{i \times a_2 : a_2 \in A_2\})$$

The products of DCPs are defined in the obvious way:

$$(E_1, G_1) \,\|\, (E_2, G_2) \;=\; (E_1 \,\|\, E_2, G_1 \times G_2)$$
$$(E_1, G_1) \rightleftharpoons (E_2, G_2) \;=\; (E_1 \rightleftharpoons E_2, G_1 \times G_2)$$

The state diagram for $\mathcal{Z}_2 \rightleftharpoons \mathcal{Z}_2$ is shown in Figure 1.

We will say that an environment or DCP is parallel (or serial) *separable* if it is isomorphic to a product of environments or DCPs.

### 5.1.1 SOLVABILITY OF SEPARABLE DCPS

The important property of separable DCPs is that their solutions can be constructed from solutions to their components:

**Lemma 1** *Let $p_1$ be a policy which solves $D_1$ and halts from all states in some set of initial states $I_1$, and let $p_2$ be a policy which solves $D_2$ and halts from all states in $I_2$. Then the policy*

$$p(x, y) = p_1(x) \times p_2(y)$$

*solves $D_1 \,\|\, D_2$ and halts from all states in $I_1 \times I_2$. (Note that here we are using the convention of treating $p$, a function over pairs, as a function over two scalars.)*

**Lemma 2** *Let $p_1$ be a policy which solves $D_1$ from all states in some set of initial states $I_1$, and let $p_2$ be a policy which solves $D_2$ from all states in $I_2$. Then any policy for which*

$$p(x, y) = p_1(x) \times i \text{ or } i \times p_2(y)$$

*and*

$$y \in G_2, x \notin G_1 \;\Rightarrow\; p(x, y) = p_1(x) \times i$$
$$x \in G_1, y \notin G_2 \;\Rightarrow\; p(x, y) = i \times p_2(y)$$

*will solve $D_1 \rightleftharpoons D_2$ and halt from all states in $I_1 \times I_2$.*

Note that the parallel and serial cases are different. One would expect the parallel case to be easier to solve because the policy can perform actions on both state components simultaneously. In fact it is more difficult because one is *required* to perform actions on both simultaneously and this leaves the agent no way of preserving one solved subproblem while solving another. Consider a "flip-flop" environment $F = (\{0, 1\}, \{flip\})$ where $flip(x) = 1 - x$. $F$ has the property that every state is accessible from every other state. $F \rightleftharpoons F$ also has this property. $F \,\|\, F$, however, does not. $F \,\|\, F$ has only one action, which flips both state components at once. Thus only two states are accessible from any given state in $F \,\|\, F$: the state itself and its flip. As with the king, the problem is fixed if we add the identity action to $F$. Then it is possible to leave one component of the product intact, while changing the other. The identity action, while sufficient, is not necessary. A weaker, but still unnecessary, condition is that $F$ have some action that always maps goal states to goal states.

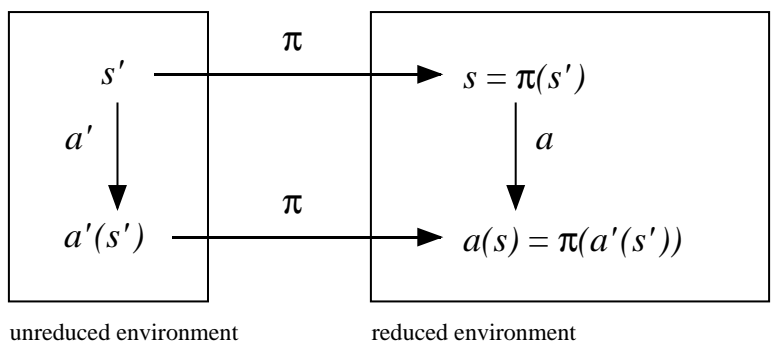

Figure 2: A simple reduction from an environment $E'$ to $E$. Here $s$ and $s'$ are corresponding states from the reduced and unreduced environments respectively and $a$ and $a'$ are corresponding actions. A projection $\pi$ is a simple reduction if it "commutes" with actions, so that $\pi(a'(s')) = a(\pi(s'))$, or alternatively, $\pi \circ a' = a \circ \pi$. Thus regardless of whether we take the projection before or after the action, we will achieve the same result.

### 5.2 Reduction

Another important kind of structure is when one environment can be considered an abstraction of another (Newell, Shaw, & Simon, 1960; Sacerdoti, 1974; Knoblock, 1989). The abstract environment retains the fundamental structure of the concrete environment but removes unimportant distinctions among states. An abstract state corresponds to a set of concrete states and abstract actions correspond to complicated sequences of concrete actions.

We will say that a projection $\pi$ from an environment $E'$ to another environment $E$ is a mapping from the state space of $E'$ to that of $E$. We will say that $\pi$ is a *simple reduction* of $E'$ to $E$ if for every action $a$ of $E$, there is a corresponding action $a'$ of $E'$ such that for any state $s'$

$$\pi(a'(s')) = a(\pi(s'))$$

or equivalently, that

$$\pi \circ a' = a \circ \pi$$

where $\circ$ is the function composition operator. We will say that $a'$ is a $\pi$-implementation of $a$ and we will use $\mathcal{A}_\pi$ to denote the function mapping $E$-actions to their implementations in $E'$.

It is possible to define a much more powerful notion of reduction in which implementations are allowed to be arbitrary policies. It requires a fair amount of additional machinery, however, including the addition of state to the agent. Since simple reduction will suffice for our purposes, we will simply assert the following lemma, which is a direct consequence of the more general reduction lemma (Horswill, 1995):

**Lemma 3** *Let $\pi$ be a simple reduction from $E'$ to some environment $E$ and let $(E', G')$ be a DCP. If a policy $p$ solves $(E, \pi(G'))$, then*

$$p_\pi = \mathcal{A}_\pi \circ p \circ \pi$$

*solves* $(E', G')$.

### 5.3 Related Work

Most formal models of environments use state-space descriptions of the environment, usually finite-state machines. Rosenschein and Kaelbling used finite state machines to represent both agent and environment (1987, 1989, 1986). Their formalization allowed specialized mechanisms to be directly synthesized from descriptions of desired behavior and a formalization of the behavior of the environment. The formalization was powerful enough to form the basis of a programming language used to program a real robot. Later, Rosenschein developed a method for synthesizing automata whose internal states had provable correlations to the state of the environment given a set of temporal logic assertions about the dynamics of the environment. Donald and Jennings (1992) use a geometric, but similar, approach for constructing virtual sensors. Lyons and Arbib (1989) model both organisms and robots using process algebras, and Beer (1995) employs the formalisms of dynamic systems theory.

Wilson (1991) has specifically proposed the classification of simulated environments based on the types of mechanisms which can operate successfully within them. Wilson also used a finite state formalization of the environment. He divided environments into three classes based on properties such as determinacy. Todd and Wilson (1993) and Todd *et al.* (1994) taxonomized grid worlds in terms of the behaviors that were successful in them. Littman (1993) used FSM models to classify environments for reinforcement learning algorithms. Littman parameterized the complexity of RL agents in terms of the amount of local storage they use and how far into the future the RL algorithm looks. He then empirically classified environments by the the minimal parameters that still allowed an optimal control policy to be learned.

There is also an extensive literature on discrete-event dynamic systems (Košecká, 1992), which also model the environment as a finite state machine, but which assume that transition information (rather than state information) is visible to the agents.

An alternative to the state-machine formalism can be found in the work of Dixon (1991). Dixon derives his semantics from first order logic, in which the world comes individuated into objects and relations, rather than on the state-space methods used here. Dixon's "open" approach also avoids the need to define the environment as a single mathematical structure. Like this work, Dixon's work attempts to formally model the assumptions a system makes about its environment. Dixon's interest, however, is on what an individual program means rather than on comparing competing programs.

### 6. TOAST

TOAST (Agre & Horswill, 1992) is a program that simulates a short-order cook in a reasonably detailed simulation of a kitchen (see Figure 3). In TOAST, the world consists of a set of objects such as ovens, pans, cutting boards, globs of pancake batter, individual eggs, and customers of the restaurant. Each object has a type (e.g., `EGG`) and all objects of a given type have a common set of possible states and a common set of possible operations that can be performed on them. An action involves a set of objects of given types. The action can require that the objects be in specified states and it may change the states of those objects, but no others. For example, the `MIX` operation would involve objects of type

| Time | Event | |
|---|---|---|
| 0 | (BREAK-EGG EGG-11 BOWL-4) | **[Making omelette]** |
| 1 | (ADD-EGG EGG-10 OMELETTE-BATTER-0) | |
| 2 | (ADD-EGG EGG-9 OMELETTE-BATTER-0) | |
| 3 | (BEAT OMELETTE-BATTER-0 WHISK) | |
| 4 | (MOVE PAN-4 BURNER-4) | |
| 5 | (MOVE BUTTER-PAT-15 PAN-4) | |
| 6 | (MELT BURNER-4 PAN-4 BUTTER-PAT-15) | |
| 7 | (MOVE SLICE-23 TOASTER) | **[Waiting for butter so making toast]** |
| 8 | (START TOASTER SLICE-23) | |
| 9 | (MOVE KNIFE-4 PLATE-1) | **[Waiting for toast so setting table]** |
| 10 | \*\*\* Done with goal (KNIFE CLEAN PLATE-1) \*\*\* | |
|    | (MOVE PLATE-1 KITCHEN-TABLE) | |
| 11 | (MOVE FORK-4 PLATE-1) | |
| 12 | \*\*\* Done with goal (FORK CLEAN PLATE-1) \*\*\* | |
|    | (MOVE SPOON-4 PLATE-1) | |
|    | Toaster pops! | |
| 13 | (MOVE BUTTER-PAT-14 KNIFE-3) | **[Back to toast]** |
| 14 | (BUTTER SLICE-23 KNIFE-3 BUTTER-PAT-14) | |
| 15 | (POUR-OMELETTE-BATTER OMELETTE-BATTER-0 …) | **[Butter melted so back to omelette]** |
| 16 | (MOVE SLICE-23 PLATE-1) | **[Setting table]** |
| 17 | \*\*\* Done with goal (SLICE BUTTERED PLATE-1) \*\*\* | |
|    | \*\*\* Done with goal (SPOON CLEAN PLATE-1) \*\*\* | |
|    | (POUR-FLOUR FLOUR BOWL-3) | **[Making pancake]** |
| 18 | (ADD-SUGAR SUGAR PANCAKE-BATTER-0) | |
| 19 | (ADD-BAKING-POWDER BAKING-POWDER PANCAKE-BATTER-0) | |
| 20 | (FOLD OMELETTE-0 SPATULA-2) | **[Tending omelette]** |
| 21 | (ADD-MILK MILK-DISPENSER PANCAKE-BATTER-0) | **[Back to pancakes]** |
| 22 | (ADD-EGG EGG-8 PANCAKE-BATTER-0) | |
| 23 | (MIX PANCAKE-BATTER-0 SPOON-3) | |
| 24 | (MOVE PAN-3 BURNER-3) | |
| 25 | (FLIP OMELETTE-0 SPATULA-2) | **[Tending omelette]** |
| 26 | (MOVE BUTTER-PAT-13 PAN-3) | **[Pancake]** |
| 27 | (MELT BURNER-3 PAN-3 BUTTER-PAT-13) | |
| 30 | (MOVE OMELETTE-0 PLATE-1) | **[Finishing omelette]** |
| 31 | \*\*\* Done with goal (OMELETTE COOKED PLATE-1) \*\*\* | |
| 36 | (SPOON-BATTER PANCAKE-BATTER-0 PAN-3 BUTTER-PAT-13) | **[Pancake]** |
| 46 | (FLIP PANCAKE-0 SPATULA-2) | |
| 56 | (MOVE PANCAKE-0 PLATE-3) | |
| 57 | \*\*\* Done with goal (PANCAKE COOKED PLATE-3) \*\*\* | |
|    | (MOVE PLATE-3 KITCHEN-TABLE) | |
| 58 | (MOVE PAN-2 BURNER-2) | **[Pancake 2]** |
| 59 | (MOVE BUTTER-PAT-12 PAN-2) | |
| 60 | (MELT BURNER-2 PAN-2 BUTTER-PAT-12) | |
| 69 | (SPOON-BATTER PANCAKE-BATTER-0 PAN-2 BUTTER-PAT-12) | |
| 79 | (FLIP PANCAKE-1 SPATULA-2) | |
| 89 | (MOVE PANCAKE-1 PLATE-2) | |
| 90 | \*\*\* Done with goal (PANCAKE COOKED PLATE-2) \*\*\* | |
|    | (MOVE PLATE-2 KITCHEN-TABLE) | |
| 91 | (CLEAN PAN-2) | **[Cleanup]** |
| 92 | (CLEAN PAN-3) | |
| 93 | (CLEAN SPOON-3) | |
| 94 | (CLEAN SPATULA-2) | |
| 95 | (CLEAN BOWL-3) | |
| 96 | (CLEAN KNIFE-3) | |
| 97 | (CLEAN PAN-4) | |
| 98 | (CLEAN WHISK) | |
| 99 | (CLEAN BOWL-4) | |
| 100 | (TURN-OFF BURNER-2) | |
| 101 | (TURN-OFF BURNER-3) | |
| 102 | (TURN-OFF BURNER-4) | |

Figure 3: Sample run of the breakfast program. The agent was given the goals of making an omelette, two pancakes, a slice of toast, and setting the table, then cleaning up. Our comments appear in square brackets.

MIXING-BOWL, BATTER, and SPOON. It would require that the spoon be in the CLEAN state and its effects would be to put the batter in the MIXED state and the spoon in the DIRTY state. Objects can perform actions, so the TOAST agent, the oven, and the customers are each modeled as objects that perform the actions of cooking, transferring heat, and making orders, respectively.

TOAST divides most of the objects in its world into two important classes (see Figure 4). Informally, *tools* are objects that (1) are not end products of cooking and (2) are easily

*Material.* Eggs. Fresh → broken → beaten → cooked.
*Material.* Butter pat. Fresh → melted.
*Material.* Milk supply. Non-empty → empty.
*Material.* Pancake batter. Has-flour → has-sugar → has-dry → has-milk → has-all → mixed.
*Material.* Pancake. Cooking → cooked-1-side → flipped → cooked → burnt.
*Material.* Bread slice. Fresh → toasted → buttered.
*Tools.* Forks, spoons, knives, spatulas, whisks. Clean → dirty, dirty → clean.
*Containers.* Bowls, plates, pans, stove burners, countertop, toaster, bread bag.
*Active objects.* Agent, stove burners, toaster.

Figure 4: Some object types in the current system.

reset to their initial states. For example, knives and spoons are used and dirtied in the process of cooking, but they are not end products of cooking and they are easily reset to their clean state by washing. *Materials* are objects that are end products of cooking but have state graphs that form linear chains. In other words, for any state of the material, there is exactly one other state that it can be brought to and there is exactly one action that can bring it there. For example, an egg being scrambled always goes through the series of states `UNBROKEN`, `BROKEN`, `BEATEN`, `COOKED`. In the `UNBROKEN` state, the only action available on an egg is `BREAK`, after which the only action available will be `BEAT`.

TOAST is given a stock of each type of object. As it runs, the customers give it goals (orders) to prepare specific dishes. The goal specifies a type of material (e.g., "EGG"). It is satisfied by putting *some* object of that type into its finished state. *Which* egg object is cooked does not matter. TOAST manages a dynamic set of these goals and opportunistically overlaps their preparation as processes finish and scarce resources, such as stove burners, become free. TOAST uses a surprisingly simple algorithm:

```
On each clock cycle of the simulator:
  Choose a material already being cooked
  Look up the action needed to advance it to the next state
  If the action requires additional tools,
    then choose objects of the proper types
  If those objects are in their reset states
    then perform the action
    else choose one of the unreset tool objects
        look up and perform its reset action
```

This algorithm is intentionally sketchy because we have implemented many versions of it which we find intuitively similar, but which have very different control structures and require very different correctness proofs. The task of the next section will be to draw out their similarities and produce a coherent theory of them.

The TOAST algorithm has two interesting features:

- Most of the algorithm proceeds by table-lookup.

- The algorithm is *stateless*: no internal plans or models are stored in the agent; all information used to choose actions is stored in the world.

Table lookup implies that the algorithm is fast and simple. Statelessness makes the algorithm simple as well, and relatively robust in the face of unexpected perturbations.

## 7. Modeling the Toast World

Why does Toast work? More specifically, what properties of its environment does it rely upon to work? In general, our strategy is to identify a series of structures in the environment that permit Toast's tasks to be factored, and then to define a series of reductions that permit more complex versions of Toast's problem to be defined in terms of simpler ones. We do not claim any vast generality for the Toast architecture; we simply observe that the environmental regularities that Toast relies upon are common to many other environments, and we suggest that our method in arguing for Toast's architecture seems likely to extend to other types of structure in the environment. Although different versions of Toast rely on different structures, we will show below that all the versions rely on:

1. The factorability of the environment into individual objects. Factoring allows us to construct solutions to problems from solutions to subproblems for the individual factors.

2. The special properties of the tool and material object classes.

3. The maintenance of *invariants* in the agent's own activity that introduce new structure into the environment.

The formalization of the properties of tools and materials is simple. The precise formalization of factorability into objects, however, is surprisingly difficult because the environment is not directly factorable using the methods we have developed so far. We will solve the problem by defining a new factoring technique called *uniform reduction*, in which the environment is viewed as a collection of overlapping instances of schematic environments, each containing the minimal set of objects necessary to perform the task. The agent solves the task by choosing one of these instances and reducing the goal in the true environment to the solution of that schematic instance. To do this, the agent must keep track of which instance it is operating on as it goes along. This could be accomplished with internal memory, of course, but then the agent would need more and more memory as it performs more and more tasks concurrently. We will show that by structuring its activity, the agent can make this information manifest in the environment, thus "storing" the information in the world.

### 7.1 Single-Material Worlds

We will start by defining the schematic environment for Toast. The environment has exactly one material to be cooked and one of each tool needed to cook it. To simplify further, we will start by ignoring even the tools. Then we will

1. Solve the no-tools case.

2. Reduce the self-resetting tools case to the no-tools case.

3. Reduce the general case to the self-resetting tools case.

### 7.1.1 Single-Material Worlds with No Tools

Since materials have linear chains as their state spaces, action in them is restricted, to say the least. In the case of an egg, we might have the chain:

$$\text{fresh} \xrightarrow{\text{break}} \text{broken} \xrightarrow{\text{beat}} \text{beaten} \xrightarrow{\text{heat}} \text{cooked} \xrightarrow{\text{heat}} \text{burnt}$$

(We will assume that the identity, or "nop," action is always available in every state. This is not a trivial assumption.) In any given state, only one non-trivial action can be executed, so action selection for an agent is trivial. When solving a DCP involving a single-material world one of the following must always hold:

- The current state is a *goal state*, so we need only execute the identity action.

- The current state is a *pregoal state*: some goal state is later in the chain than the current state, so we can reach it by executing the unique action that brings us to the next state in the chain.

- The current state is a *postgoal state*: all goal states are earlier in the chain, so the problem is unsolvable.

All that really matters in single-material worlds, therefore, is how many states there are and in which direction the goal lies relative to the current state. In a sense, there is only really one single-material world, or rather one class of them, namely the chains $\mathcal{C}_n$ of given length:

$$\mathcal{C}_n = (\{1, ..., n\}, \{inc_n, i\})$$

(Note this is just the same as the environment $\mathcal{Z}_n$, but without the actions that move backward along the chain.)

**Proposition 1** *All single-material worlds of $n$ states are reducible to $\mathcal{C}_n$*

*Proof:* Let $E = (S, A)$ be the single-material environment. Define $\pi \colon S \to \{1, ..., n\}$ by letting $\pi(s)$ be $s$'s position in $E$'s state chain, *i.e.* the first state maps to 1, the second to 2, etc. Let $action(s)$ denote the unique action that can be performed in state $s$. Then

$$p_{inc_n}(s) = (action(s))(s)$$

is a $\pi$-implementation of $inc_n$ and so $E$ is reduced. $\square$

Just as there is only one real class of single-material worlds, there is only one real class of policies for single-material DCPs:

$$p_{\mathcal{C}_n, G}(s) = \begin{cases} i, & \text{if } s \in G \\ inc_n, & \text{otherwise} \end{cases}$$

which clearly solves the DCP $(\mathcal{C}_n, G)$ for any $n$ and valid $G$.

**Corollary 1** *If a goal $G$ is solvable in a single-material environment $E$ with no tools, then it is solved by the policy*

$$p_{E,G}(s) = \begin{cases} i, & \text{if } s \in G \\ (action(s))(s), & \text{otherwise} \end{cases}$$

### 7.1.2 Single-Material Worlds with Single-State Tools

Now suppose the world contains a material and a set of tools, but those tools always clean or otherwise reset themselves after use. Self-resetting tools have only one state, and so they are a trivial kind of environment. We define the "singleton" environment as the environment with exactly one state:

$$\mathcal{S} = (\{\text{ready}\}, \{i\})$$

All single-state environments are isomorphic to $\mathcal{S}$, so we model an environment consisting of a material $M = (S, A)$ and a self-resetting tool as $M \parallel \mathcal{S}$. Its state space is simply $S \times \{\text{ready}\}$ and its actions are just the set

$$\{a' \colon (s_M, \text{ready}) \mapsto (a(s_M), \text{ready}) | a \in A\}$$

Each such action performs some action from $M$ on the $M$-component of the product's state and leaves the $\mathcal{S}$ component unchanged. By induction, we have that:

**Proposition 2** *Any environment $M$ is isomorphic to $M \parallel \mathcal{S}^n$.*

And so single-state-tool worlds are trivially reducible to tool-free worlds.

### 7.1.3 Single-Material Worlds with General Tools

The general tool environment is identical to the single-state tool environment, except that actions can change the states of tools in addition to the states of materials. We can solve the general tool case using a solution to the single-state tool case by resetting tools whenever they are dirtied.

The proof is simple, but requires that we formalize the notion of being a tool. Let $E$ be an environment with a state space of the form $S_1 \times S_2 \times ... \times S_n$. Let $a$ be an action of $E$ and $S_i$ be a component of its state space. We will say

- $a$ is *independent of* $S_i$ if $a$ never changes $S_i$ and it has the same result regardless of the value of $S_i$.

- $a$ is *focused on* a component $S_i$ if it is independent of all other components.

- $S_i$ *is a tool* if it has some privileged value $ready_i \in S_i$ such that:

    - From any state $(s_1, ..., s_i, ..., s_n)$ of $E$, we can reach the state $(s_1, ..., ready_i, ..., s_n)$ using only actions focused on $S_i$.
    - For any action $a$, $a$ is either independent of the $S_i$, focused on $S_i$, or else only defined on states whose $S_i$ component is $ready_i$.

Now we can prove the general tool case is reducible to the single-state tool case:

**Lemma 4** *Any environment with tool components can be reduced to one in which the tools have been replaced by singletons. Specifically, let $D = ((S, A), G)$ be a DCP and let $ready_T \in T$, and $A' = \{a' \colon (s, ready_T) \mapsto (a(s), t) | a \in A, s \in S, \text{and } t \in T\}$. Then $D' = ((S \times T, A' \cup A_t), G \times \{ready_T\})$ is reducible to $D$ when $T$ is a tool in $D'$.*

*Proof:* Let $p_D$ be a solution (policy) for $D$. By the definition of a tool, there must be a policy $p_T$ that will bring $D'$ from any state $(s, t)$ to $(s, ready_t)$ without changing the $S$ component. Let $\pi$ be the projection from $D'$ to $D$ given by

$$\pi(s, t) = \begin{cases} s, & \text{if } t = ready_T \\ \bot, & \text{otherwise} \end{cases}$$

For each $a \in A$, we define the $\pi$-implementation of $a$, $p_a$ by

$$p_a(s, t) = \begin{cases} a', & \text{if } t = ready_T \\ p_T, & \text{otherwise} \end{cases}$$

and so $D'$ is reducible to $D$. The general case of multiple tools follows from induction. □

## 7.2 Multiple-Material Worlds with Single-Material Goals

To reprise: we want to factor the environment into its individual objects and then describe TOAST as a composite of techniques for operating on the individual factors. We cannot properly define environments as Cartesian products of individual objects defined in isolation because we have no way of expressing actions involving multiple objects. We *can*, however, define a set of objects in the context of a minimal, schematic environment containing one copy of each object. Having done so, we now want to recapture the notion of an environment being some kind of product of objects of different types. We will do this by showing that an environment with two eggs can be thought of as two overlapping copies of an environment with one egg; the copies differ only in the choice of the egg.

We will treat environments as having state spaces formed as products of the state spaces of their objects. A state of the environment is a tuple of the states of its objects. A *binding* of the schematic environment to the real environment is a particular kind of projection from the complex environment to the schematic, one which is also a reduction. If all reasonable projections are valid bindings, then we will say the environment is *uniformly reducible* to the schematic environment.

### 7.2.1 BINDINGS AND UNIFORM REDUCIBILITY

Let $E'$ and $E$ be environments with state spaces built as Cartesian products of a family of disjoint sets $\{S_i\}$. The $S_i$ might represent the state spaces of object types like egg and fork. $E'$ and $E$ would then each have state spaces make of up some number of copies of egg and fork.

We will say that a projection $\pi$ from $E'$ to $E$ is *simple* if every component of its result is a component of its argument. That is

$$\pi(s_1, s_2, s_3, ..., s_n) = (s_{i_1}, s_{i_2}, ..., s_{i_m})$$

for some $i_1, ..., i_m$ in $[1, n]$. Thus $\pi$ takes a $E'$-state, $s'$, probably throws away some of its components, and possibly rearranges the rest to form a new tuple. For example, $\pi$ might single out a particular egg's state and/or a particular fork's state and throw the other state components away. When a projection is simple, we can define a kind of inverse for it, which we will call its *back-projection*. We will define the back-projection, $\pi^-(s, s')$, of $\pi$ to be the function whose result is $s'$ with those components that $\pi$ keeps replaced by their corresponding components from $s$. For example, if $\pi$ is defined by

$$\pi(s'_1, s'_2, s'_3) = (s'_3, s'_2)$$

then its back-projection would be given by:

$$\pi^-((s_a, s_b), (s'_1, s'_2, s'_3)) = (s'_1, s_b, s_a)$$

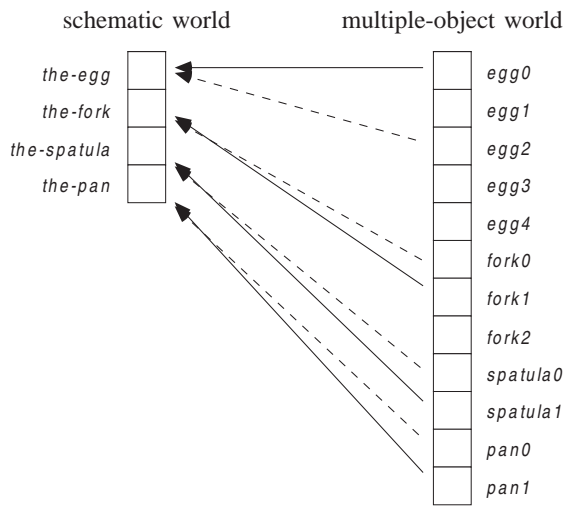

Figure 5: A binding (solid vectors) and an alternate binding (dashed).

We will say that a simple projection is a *binding* of $E$ to $E'$ if it is also a simple reduction of $E'$ to $E$ (see Figure 5).

**Lemma 5** *Let $\pi$ be a binding of $E$ to $E'$. Then $\mathcal{A}_\pi$ is given by*

$$\mathcal{A}_\pi(a) = a_\pi, \text{ where } a_\pi(s') = \pi^-(a(\pi(s')), s')$$

*That is, the implementation of an E-action is simply $\pi$'s back-projection composed with that action and $\pi$.*

The proof follows from the definitions of simple projection and back-projection. We will say that $E'$ is *uniformly reducible* to $E$ if every simple projection from $E'$ to $E$ is a binding.

7.2.2 Existential Goals

Toast is given the goal of putting *some* instance of a given material in its finished state. We will call this an *existential goal* because it is satisfied by exactly those environment states in which there exists an object of a specified type in a specified state. Let $(E, G)$ be a DCP and let $E'$ be uniformly reducible to $E$. We define the *existential goal* $\exists_{E,E'} G$ of $G$ in $E'$ to be the set of states in $E'$ that project under some binding to a goal state in $(E, G)$:

$$\exists_{E,E'} G = \bigcup_{\pi \text{ a binding of } E \text{ to } E'} \pi^{-1}(G)$$

where $\pi^{-1}(G) = \{s': \pi(s') \in G\}$ is the set of states that map to goal states under $\pi$. Given a solution to a schematic goal in a schematic environment, we can easily construct any number of solutions to the existential goal:

**Lemma 6** *If policy p is a solution for the problem $(E, G)$ from initial states $I$, and $\pi$ is a binding from $E'$ to $E$, then*

$$p_\pi = \mathcal{A}_\pi \circ p \circ \pi$$

*is a solution for $(E', \exists_{E,E'} G)$ from initial states $\pi^{-1}(I)$, where $\mathcal{A}_\pi$ is the function mapping actions in $E$ to their corresponding actions under $\pi$ in $E'$.*

The TOAST algorithm implements a policy which is a composition of a schematic solution and a binding that maps it onto the real world. Consider the problem of cooking an egg. The schematic solution might be:

> **break** *the-egg* into *the-pan*
> **beat** *the-egg* in *the-pan* using *the-whisk*
> **heat** *the-egg* in *the-pan*

here the boldface verbs **break**, **beat**, and **heat** name actions. The italicized expressions *the-egg* and *the-pan* name the objects (state components) that they affect in the simplified world. The binding then determines objects in the real world to which those state components correspond. Given a binding, the main control structure need only remember the sequence **break**, **beat**, **heat**. Each of these may have preconditions on the states of the tools (i.e. the whisk needs to be clean), but they can be handled by reduction given policies for resetting the tools.

### 7.2.3 BINDING MAPS

Given the basic policy for cooking a single egg with a single pan and whisk, we can construct a policy to achieve the goal by composing the basic policy with a binding. This policy will solve the goal from any state in which the bound material is in a non-postgoal state. For a policy to solve the goal from *any* solvable state, it must be able to change bindings at run time. We will call a function from states to bindings a *binding map*.

One simple policy for choosing bindings is to impose some *a priori* ordering on the objects and always use the first acceptable object in the ordering. The ordering might be random, or it might correspond to order imposed by a visual search mechanism. From a formal standpoint, the ordering does not matter, so we can, without loss of generality, use the left-to-right order of state components in the environment's state tuple. Let $M_0$ be some binding map that always chooses the leftmost pregoal material and uses some fixed mapping for the tools (we do not care what). This mapping allows us to construct a true solution, and one that requires no internal state in the agent:

**Proposition 3** *The policy*

$$p_{M_0}(s) = (\mathcal{A}_{M_0(s)} \circ p \circ (M_0(s)))(s)$$

*is a solution from any state for which $M_0$ is defined.*

*Proof:* By assumption, $M_0$ is defined in the initial state. The environment must then map to a solvable state under $M_0$ in the initial state. Since $p$ is, by assumption, a solution for the problem in $E$, $p_{M_0}$ must solve the problem in $E'$ unless $M_0$ changes value before the $p_{M_0}$ can solve the problem. Suppose it does. Then the environment must go from a state $s'_0$, in which some state component of $E'$ is the leftmost pregoal material, to a state $s'_1$, in which some other component is the leftmost pregoal material. This can only happen if (a) the leftmost pregoal material in $s'_0$ is changed to be in a goal state in $s'_1$ or (b) some other component that was not pregoal in $s'_0$ becomes pregoal in $s'_0$. Case (b) is impossible and case (a) implies that $s'_1$ is itself a goal state. Thus $p_{M_0}$ must be a solution. □

### 7.3 Multiple Goals: Metabolism

Thus far, we have not considered what happens once a policy achieves its goal. Since agents rarely set out to achieve a goal and die, we now want to consider how to account for extended activity involving many goals.

One important class of extended activities is when an agent transforms a whole class of identical objects. We will call this *metabolizing* the class. Metabolism can be useful or it can make extra work: cooking 100 eggs is useful, at least if you are feeding a lot of people; dirtying 100 forks, however, probably means you have to wash them all.

Whether a policy metabolizes an object class depends in large part on the binding map it uses. The policy $p_{M_0}$ metabolizes its materials because the material being worked on ceases to be the leftmost pregoal material as soon as it arrives in a goal state. When this happens, $M_0$ changes bindings and the agent starts to work on a different object. Policy $p$ never actually sees a material in a goal state. Of course, the property of being "leftmost" is an artifact of our formalism. What matters to the property of metabolism is simply that the binding map implement some ordering on the instances of the material and always choose the minimum under that ordering of the objects that are in pre-goal states. Such an ordering might be implemented by the agent visually scanning its work surface for an uncooked egg, but always scanning left-to-right and top-to-bottom. We will return to these issues in section 8.

Other binding maps lead to other kinds of behavior, some of which are pathological. If the binding map always chooses the same binding, then metabolism ceases. If the binding map always chooses uncooked eggs but doesn't impose any ordering on them, it might start cooking an infinite number of eggs without ever actually finishing any one of them.

Metabolism is also an issue for tool use. To metabolize its materials, $p_{M_0}$ must repeatedly reset its tools. An alternate policy is to metabolize the tools too. Let us define $M_1$ to be the binding map that uses not only the leftmost pregoal material but also the leftmost reset tools. Then clearly,

$$p_{M_1}(s) = (\mathcal{A}_{M_1(s)} \circ p \circ (M_1(s)))(s)$$

is a solution from any state for which $M_1$ is defined. This policy treats tools as *disposable*. So long as there is an infinite supply of fresh tools, $p$ will see a succession of states in which tools are in their reset states. It will never need to execute a resetting action and so the environment is effectively a single-state-tool environment. Thus the reduction of section 7.1.3 is unnecessary.

### 7.4 Multiple Goals: Interleaved Execution

Metabolism involves performing the same transformation uniformly to instances of the same type of object: cooking all the eggs, or cleaning/dirtying all the forks. Often times, however, an agent will work toward different kinds of goals at once. This can often be done by interleaving the actions from solutions to the individual goals. We will say that an *interleaving* $I$ is a function that returns one or the other of its first two arguments, depending on a third state argument:

$$I(s, p_1, p_2) \in \{p_1, p_2\}, \text{ for all } s$$

When the last two arguments of $I$ are policies, the result is itself a policy, so we will define the notation:

$$I_{p_1, p_2}(s) = (I(s, p_1, p_2))(s)$$

If we wanted to simultaneously make toast and cook an egg, then a good interleaving of a toast-making policy and an egg-cooking policy would be one that chose the egg-making policy whenever the egg had finished its current cooking step (and so was ready to be flipped or removed from the pan) and chose the toast-making policy when the egg was busy cooking. A bad interleaving would be one that always chose the toast-making policy.

An interleaving $I$ is *fair for* $p_1$ and $p_2$ if starting from any state, $I_{p_1,p_2}$ will after some finite number of steps have executed both $p_1$ and $p_2$ at least once. Finally, we will say that two bindings are *independent* if they map disjoint sets of components to their images. Binding independence is a special case of subgoal independence: two policies can't possibly interfere if they alter distinct state components. Fairness and binding independence are sufficient conditions for an interleaving to solve a conjunctive goal:

**Lemma 7** *Let* $p_1 = \mathcal{A}_{\pi_1} \circ p_1' \circ \pi_1$ *and* $p_2 = \mathcal{A}_{\pi_2} \circ p_2' \circ \pi_2$ *be policies that solve the goals* $G_1$ *and* $G_2$, *respectively, and halt. If* $\pi_1$ *and* $\pi_2$ *are independent and* $I$ *is a fair interleaving for* $p_1$ *and* $p_2$ *then* $I_{p_1,p_2}$ *solves* $G_1 \cap G_2$ *and halts.*

*Proof:* Since $I$ is a fair interleaving, each of the two policies will be executed in finite time, regardless of starting state. By induction, for any $n$, there is a number of steps after which $I$ is guaranteed to have executed at least $n$ steps of each policy.

The policy $p_1$ is the composition of a policy $p_1'$ for a state space $S_1$ with a binding. If $p_1$ solves $G_1$ and halts, then it must do so by having $p_1'$ solve $\pi(G_1)$ and halt in some finite number of steps $n$. During execution, the environment goes through a series of states

$$s_0, s_1, ..., s_n$$

which project under $\pi_1$ to a series of states

$$s_0', s_1', ..., s_n'$$

we claim that any execution of the interleaving $I_{p_1,p_2}$ must bring the environment through a sequence of states that project under $\pi_1$ to

$$(s_0')^+, (s_1')^+, ..., (s_n')^+, ...$$

that is, a string of states in which $s'_0$ appears at least once, then $s'_1$, appears at least once, and so on. The only state transitions that appear are from some $s'_i$ to itself or to $s'_{i+1}$. Suppose it were otherwise. Then there must be a point at which this series is broken:

$$(s'_0)^+, (s'_1)^+, ..., (s'_i)^+, s$$

where $s$ is neither $s'_i$ nor $s'_{i+1}$. We have two cases. *Case 1:* $p_1$ executed the transition. Then we have that $p'_1(s'_i) = s \neq s'_{i+1}$, a contradiction. *Case 2:* $p_2$ executed the transition. Then $p_2$ has changed one of the state components mapped by $\pi_1$ and so $\pi_2$ and $\pi_1$ are not independent, a contradiction. Thus the interleaving solves $G_1$. By the same reasoning, it must halt in $G_1$, since $p_1$ halts in $G_1$. Also by the same reasoning, it must solve $G_2$ and halt, and hence, must solve the intersection and halt. □

A useful corollary to this is that when the same policy is applied under two independent bindings, the bindings can be safely interleaved, that is, interleaving commutes with binding:

**Corollary 2** *If $p_1 = \mathcal{A}_{\pi_1} \circ p \circ \pi_1$ and $p_2 = \mathcal{A}_{\pi_2} \circ p \circ \pi_2$ be policies that solve the goals $G_1$ and $G_2$, respectively, and halt, and $I$ is a fair interleaving for $p_1$ and $p_2$, then $\mathcal{A}_{I_{\pi_1,\pi_2}} \circ p \circ I_{\pi_1,\pi_2}$ solves $G_1 \cap G_2$ and halts.*

## 8. Implementing Policies and Bindings

We have modeled TOAST's behavior as a composition of various bindings and interleavings with a basic policy for a schematic environment. In the case of TOAST, the basic policy is simple enough to be implemented by table-lookup. The hard part is implementing the bindings and interleavings given realistic limitations on short-term memory and perceptual bandwidth.

One approach would be to assume a relatively complete representation of the world. Each egg would be represented by a logical constant and its state would be represented by a set of propositions involving that constant. A binding would be implemented as a frame structure or a set of variables that point at the logical constants. The problem is that this approach presupposes the underlying perceptual and motor systems maintain a correspondence between logical constants and eggs in the world. When one of the eggs changes, the visual system has to know to be looking at it and to update the assertions about the egg in the model.

This is not an assumption to be taken lightly. The capacity of the human perceptual system to keep track of objects in the world is extremely limited. Ballard et al. (1995)found that their experimental subjects adopt strategies that minimized the amount of world state they needed to track internally, preferring to rescan the environment when information is needed rather than memorize it in advance. The environment could even be modified during saccadic eye movements without the subjects noticing.

An alternative is to treat the limitations of the body, its locality in space, and its limited attentional and motor resources as a resource for implementing bindings directly. A person can visually focus on one object, stand in one place, and grasp at most a few objects at any one time. The orientation of the body's parts relative to the environment can be used to encode the selection of objects being operated on at the moment. In other words, it can implement a binding. Actions of the body, gaze shifts, and movements to new places can be used to shift that binding.

Another alternative is to use the states and relationships of objects in the world to keep track of bindings. An egg is being cooked if it is in the frying pan. A fork is available for use if it is in the drawer, but not if it is in the sink waiting to be washed.

In this section, we will model the use of the body and conventions to implement bindings and interleavings. To simplify the presentation and to be concrete, we will focus on materials, particularly eggs.

### 8.1 Binding, Deixis, and Gaze

To a first approximation, people can only visually recognize objects at which they are directly looking. People achieve the illusion of direct access to arbitrary objects by rapidly changing their gaze direction. Thus in addition to the normal state of the environment, our lived world contains an additional state component, our gaze direction. Since we can normally change our gaze direction without changing the world, and vice versa, our lived world $E'$ can be separated into a parallel product of the objective environment and our gaze direction:

$$E' = E \parallel D$$

Our access to this world is through our gaze, which allows us to focus in on one particular object at a time. Our gaze implements a binding, or more precisely, a binding map, since it depends on the direction of gaze. If we model gaze direction as a number indicating which object is presently foveated, we have that:

$$\pi_{gaze}(s_1, s_2, ..., s_n, d) = s_d$$

A person could implement a single-object binding just by fixating the object they wish to bind. First they would set the $D$ component to some egg, and then use $\pi_D$ as their binding. Since $\pi_D$ is really a binding map, however, rather than a true binding, the agent must pervasively structure its activity so as to ensure that its gaze need never be redirected.

### 8.2 Binding and Convention

In general, agents must maintain bindings through some sort of convention, whether it is the structuring of their internal memory, as in the case of a problem solver, or the structuring of their activity. In the case of gaze above, the agent maintains the binding through a convention about the spatial relation between its eye and the object it is binding. All versions of TOAST to date have maintained bindings using conventions about (simulated) spatial arrangement or the states of objects.

One reason TOAST cannot rely solely on gaze binding is that the technique breaks down when binding multiple objects. The agent must continually move its gaze among the objects of interest and so some additional convention must be introduced to ensure that when its gaze leaves the egg and later returns, it always returns to the *same* egg. (This assumes, of course, that TOAST must return to the same egg. In some tasks it may suffice for TOAST to return to some functionally equivalent egg. If it is preparing three fried eggs and its attention is distracted while preparing to break the second one, it is alright if its attention returns to the third egg, so long as it gets back to the second egg eventually.)

*State conventions*

The original version of TOAST used the convention *eggs were bound to a cooking task iff they were not in their starting (unbroken) state.* Eggs were therefore bound using the binding map

$$\pi_{Toast}(s) = \text{ the state of the unique egg in } s \text{ that is in an unbroken state}$$

which the agent can implement by first visually searching for an unbroken egg, and then using $\pi_{gaze}$. By corollary 2, the interleaving of the cooking of multiple eggs can be accomplished by interleaving the bindings of the eggs. For example, we might assume that the visual system searched non-deterministically or in a round-robin fashion for eggs. Any fair interleaving will suffice.

*Spatial conventions*

Later on in our development of TOAST, we found it useful to adopt the convention that *eggs were bound to a cooking task iff they were located in a designated workspace.* Cooking eggs are on the counter or in the frying pan, while idle eggs are in the refrigerator. This convention lets the agent use space as an external memory for binding information. To bind the egg, the agent faces the workspace and performs visual search for an egg. Any egg it finds will be an egg being cooked, since idle eggs will be out of view.

This still leaves open the issue of fairness. An extreme but elegant solution to the fairness problem is to use multiple workspaces and employ the convention that each workspace defines a unique binding. To cook two eggs, the agent just works on cooking whatever egg is in front of it, but it spins in place so that it alternates between workspaces.

Formally, the environment then consists of two copies of the workspace and the objects therein plus an extra state component that determines which workspace the agent faces. The agent's perceptual system implements a binding map in which one or the other of the two workspaces is bound depending on the agent's orientation. Given a policy for cooking one egg in one workspace, we can construct a policy for cooking two eggs in two by interleaving the policy with a "flipping" operation that switches the workspaces:

**Proposition 4** *Let $E = (S, A)$ be an environment, $p$ be a policy that solves some goal $G$ in $E$ and halts, and let $D$ be an environment with two states, 0 and 1, and two actions, $i$ (the identity) and $flip$ which moves the environment into the opposite state from its present state. Consider the product environment:*

$$E' = E \rightleftharpoons E \rightleftharpoons D$$

*and the binding map from $E'$ to $E$:*

$$M_D(s_0, s_1, d) = s_d$$

*Any fair interleaving of the policies:*

$$p_{M_D} = \mathcal{A}_{M_D} \circ p \circ M_D$$

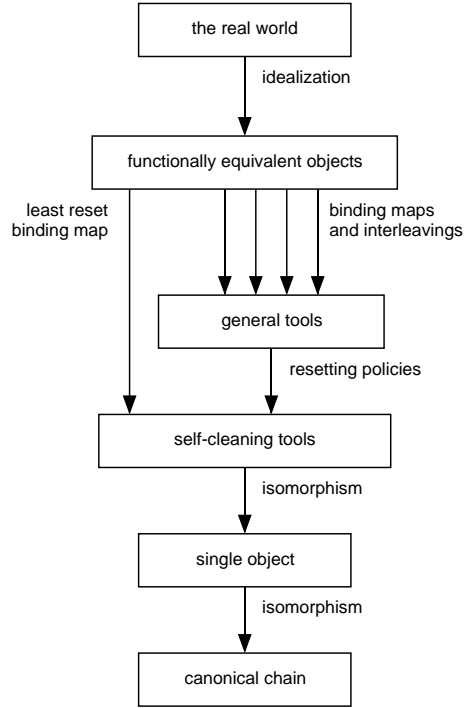

Figure 6: Various alternative reductions used in TOAST.

and

$$p_{flip}(s_0, s_1, d) = i \times flip$$

is a solution to the problem $(E', (G \times G \times \{0, 1\}))$.

*Proof:* Consider the bindings $\pi_0 \colon (s_0, s_1, d) \to s_0$ and $\pi_1 \colon (s_0, s_1, d) \to s_1$, and let $p_{\pi_0} = \mathcal{A}_{\pi_0} \circ p \circ \pi_0$ and $p_{\pi_1} = \mathcal{A}_{\pi_1} \circ p \circ \pi_1$. Since the binding map $M_D$ alternates between the bindings $\pi_0$ and $\pi_1$, any fair interleaving of $p_{M_D}$ with $p_{flip}$ is equivalent to some interleaving of $p_{\pi_0}$, $p_{\pi_1}$ and $p_{flip}$. We would like to show that this interleaving is also fair, that is, that each of $p_{\pi_0}$ and $p_{\pi_1}$ will get run in finite time. We can see this from the fact that with each execution of $p_{flip}$ switches $M_D$ from one binding to another. An objection is that this leaves open the possibility that that $p_{flip}$ will always get run twice in a row, thus returning the environment to its original state and so preventing $M_D$ from switching bindings. This cannot occur, however, since it would introduce a loop, causing the interleaving to run $p_{flip}$ forever, never running $p_{M_D}$, and so violating the assumption of fairness of the interleaving of $p_{M_D}$ and $p_{flip}$. Thus the interleaving of $p_{\pi_0}$, $p_{\pi_1}$ and $p_{flip}$ must be fair. Now note that $p_{\pi_0}$ solves the goal $G \times S \times \{0, 1\}$ and halts, $p_{\pi_1}$ solves the goal $S \times G \times \{0, 1\}$ and halts, and $p_{flip}$ solves the goal $G \times G \times \{0, 1\}$ and halts. Thus by lemma 7, the interleaving solves the intersection of these goals, which is $G \times G \times \{0, 1\}$. □

## 9. Reductions and the Structure of Toast

We have shown how the cooking problem can be solved by a series of reductions and conventions. Binding allows the reduction of the problem to a schematic world in which action is greatly restricted and so action selection is greatly simplified. This world can be further reduced, given algorithms for resetting tools, to a world in which tools are always reset. This world, in turn, is equivalent to a world in which there is only one object, the material being cooked, and only one action can be taken at any given time. Such actions can be found by table lookup.

Multiple materials can be cooked by interleaving the execution of processes for cooking the individual materials. Interleaving the processes is equivalent, however, to interleaving the bindings, so the schematic-world algorithm need not even be aware that it is pursuing multiple goals. If tool bindings are continuously changed as tools are dirtied then tools are effectively disposable, tools effectively have only a single state, and the separate reduction from general tools to single-state tools is unnecessary. Material bindings can be maintained by any number of conventions involving the states and/or positions of objects.

In short, we can describe a Toast algorithm as a path through a network of possible simplifications of the problem (see Figure 6) in which every path from the actual world to the idealized single-object world defines a possible (and correct) version of the Toast algorithm.

## 10. Cognitive Autopoiesis

In formalizing our ideas about binding and gaze, we have been moving toward a theory of intentionality that depends on the agent's embedding in its world, rather than solely upon its internal models of that world. An agent can keep track of particular objects in terms of their functional significance – the roles that they play in the ongoing activity. And it can keep track of the tools and materials associated with different tasks by keeping them in different locations, for example different regions of a countertop. So far, however, our ideas on the subject have been limited to very simple cases, for example an agent switching its visual focus back and forth between two objects. To model the more complex patterns that are found in everyday life, we need a much better theory of the world in which we are embedded. This theory is partially a matter of biology and physics, of course, but it is also a matter of cultural practices for organizing activities in space. In this section, we would like to sketch a more general theory of these matters using the concept of "cognitive autopoiesis."

For Maturana and Varela (1988), autopoiesis refers to the processes by which organisms act on their environments in order to provide the conditions for their own continued functioning. Cognitive autopoiesis refers to the active means by which agents structure their environments in order to provide the conditions for the own cognitive activities. These include most basically the means by which agents provide for the factorability of environments: engaging in customary activities, using the customary tools and materials for them, partitioning the activities in the customary ways, and so on. But it also includes a range of more subtle phenomena. Kirsh (1995), for example, has drawn the useful distinction between actions that aim at achieving functional goals (beating eggs, sweeping floors) and actions that aim at facilitating cognition (setting out the right number of eggs at the beginning, opening the curtains so that dust will be more visible). Actions can, of course, serve both purposes, for example when one chooses to boil water in a kettle rather than a saucepan: each strategy achieves the result, but the latter will also provide a sign that it is possible to take the next action, for example preparing tea. Stabilization actions (Hammond et al., 1995) also provide the cognitive conditions for other actions. One might, for example, develop the habit of leaving items by the door the moment one realizes that they need to be taken in to work.

These phenomena help in understanding what is inadequate about the concept of "the environment." If one conceptualizes "the environment" as a monolithic whole, perhaps the way it looks when viewed from an airplane, or else the way it looks when understood through the peephole of a momentary vector of sense-perceptions, it begins to seem arbitrary, chaotic, or hostile. In a certain sense it seems static, as if it has an anatomy but no physiology. But in fact the phenomena of cognitive autopoiesis reveal that the lifeworld has a great deal of living structure, and that this structure is actively maintained by agents while also providing crucial preconditions for their own cognition. Indeed it is hard to draw a clear line around an agent's cognition; if we trace the sequence of causal events that led a given agent to pour a pitcher of milk on a particular moment, this sequence will lead back and forth between the agent and its customary surroundings. It is almost as if these surroundings were an extension of one's mind.

Cognitive autopoiesis is a complex and multifaceted phenomenon and no single theory will suffice to explain it. One useful way to think about cognitive autopoiesis is spatially, in terms of a series of buffer zones between the embodied agent and the putative dangers and complexities of "the environment." For people whose lives are similar to our own, these buffer zones can be conveniently sorted under six headings:

- *The body itself*: its posture, its markings, things that might be attached to it or hung from it, prostheses, artificial markings, the things one is holding in one's hands, and so on. All of these things can serve as forms of memory, for example as a way to remember what activity one was in the middle of before some momentary distraction. The body's motility also makes possible a wide range of voluntary reconfigurations of one's physical relationship to things, for example to get a better view or better leverage.

- *Clothing*, including pockets, purses, money belts, hats, and so on. Everyone carries around various objects in ways that draw on customary practices and artifacts (cash in wallets, keys in pockets, watch on wrist, etc) while configuring these things in an evolving personal way (keys in left pocket and money in right, tissues in the hip pocket of one's coat, spare change in the outer flap of the backpack, and so on).

- *Temporary workspaces* that one occupies to perform a particular activity over a bounded period. In repairing a bicycle, for example, one might spread tools and bicycle parts about on the floor in patterns that have a cognitive significance in relationship to one's own body and cognitive and other states (Chapman & Agre, 1986). One is not claiming this space as a permanent colony (it might be located on a patio in a

   public park, for example), but one does lay claim to the space long enough to perform
   a customarily bounded task.

- *One's own private spaces*: home, desk, office, car, trunks of stuff kept in someone else's
  attic, and so forth. These spaces serve numerous functions, of course, but among these
  are the cognitive functions of providing stable locations over long periods of time for
  tools and materials, storage places for stuff that needs to be kept in adequate supply,
  practices for regulating other people's access to the stuff, and so on. These stable
  conditions are actively maintained and provide the background for a wide variety of
  more transient activities.

- *Spaces that are shared* with other people within stable, time-extended relationships.
  These spaces include living rooms, kitchens, shared office spaces, and so forth. The
  line between the private and shared spaces clearly depends on the particular culture
  and set of relationships, and the distinction might not be clear. The point is that
  the cognitive functions of the spaces are maintained through shared practices such as
  letting someone know when you borrow their stuff.

- *Public spaces* and the whole range of customary artifacts and practices that regulate activities in them. Public spaces offer fewer guarantees than private and shared
  spaces, but they do include a wide variety of supports to cognition, including signs
  and architectural conventions. It is also possible to use one's body and clothing to
  carry artifacts that provide cognitive support for dealing with public spaces.

These buffer zones do not always offer perfect protection from harm or complete support for the pursuit of goals. Shared and public spaces can be sites of conflict, for example, and these conflicts can include involuntary disruption or destruction of one's body and the other buffer zones that are customarily under one's own private control. A serious theory of activity must include an account of these phenomena as well, which are usually just as orderly in their own way as anything else.

   In any event, the nested buffer zones of ordinary life participate in a large metabolism that continually interweaves cognitive and functional purposes. Among these purposes is learning. Just as the adaptation of body parts and tools to customary activities helps channel action in customary directions, so does the existing background of objects, spaces, and practices help channel the actions of children and other newcomers in customary directions on a larger scale. Caretakers regularly construct customized types of buffer zones around the young, for example, so that it is difficult or impossible for them to get into anything that could cause harm. The lifeworld of a child, for example, differs from that of an adult who can reach up to the cookie jar and into the locked cupboard where the roach spray is kept. A growing literature has investigated the processes of cognitive apprenticeship (Rogoff, 1990), situated learning (Lave & Wenger, 1991), distributed cognition (Hutchins, 1995; Salomon, 1993), and shared construction of activities (Griffin & Cole, 1989) that go on in these systematically restrictive and supportive lifeworlds.

## 11. Conclusion

In this paper we have explored some of the ways in which the structure of the lifeworld supports agents' cognition, and we have suggested how this analysis might be expanded to cover a wider range of phenomena. Much work obviously remains to be done. Perhaps the most significant part of this work concerns a fundamental assumption of lifeworld analysis: that people use objects in customary ways. This is a plausible enough first approximation, but it is not always true. Faced with a difficulty that goes beyond the capacities of the usual practices and the artifacts that are readily available, people frequently improvise. The handle of a spoon might be used to pry open a lid, a pen might be used to fish acorns out of an exhaust duct, a book might be used to provide backing for a sheet of paper one is writing on, or a protruding section of a car's bumper might be bent straight by deliberately driving the car into a concrete wall. In these cases the underlying physical affordances of an object "show through" beyond their ready-to-hand appropriation in routine patterns of interaction. These underlying affordances can also show through in situations of breakdown, for example when a tool breaks or proves inadequate to a job. In such cases, people confer improvised meanings upon artifacts. Such phenomena are particularly important in conversation, in which each utterance is interpreted in the context created by previous utterances, while simultaneously helping to create the context for interpretation of successive utterances as well (Edwards & Mercer, 1987; Atkinson & Heritage, 1984). The point is not that the lifeworld does not exist, but rather that it is something actively created *as well as* something adapted to through socialization. One challenge for future research is to learn how computational methods might help in modeling such phenomena—and how such phenomena might help us to rethink basic ideas about computation.

## Acknowledgements

We appreciate the detailed comments of the referees. This work was funded in part by the National Science Foundation under grant number IRI–9625041. The Institute for the Learning Sciences was established in 1989 with the support of Anderson Consulting, part of the Arthur Anderson Worldwide Organization.

## Glossary of Terms

**Binding**. A simple projection (mapping between state-space components of two environments) that acts as a reduction from one environment to another (see section 7.2.1).
**Binding map**. A mapping from environment states to bindings (see section 7.2.3).
**Cartesian product**. *For sets:* $A \times B$ is the set of all pairs $(a, b)$ for $a \in A$, $b \in B$. *For environments:* an environment is the Cartesian product of two other environments iff its state space is the Cartesian product of their state spaces. Since the set of actions is left open in this definition, there are many possible ways of forming products, *e.g.* serial product, parallel product, uniform extension, *etc.*(see section 5.1).
**Discrete control problem (DCP)**. An environment and a set of goal states within it (see section 5).
**Environment**. A state machine, *i.e.* , a set of possible states and a set of possible actions mapping states to states. The sets of states and actions need not be finite (see section 5).

**Focus**. An action is focused on a state component if it only alters that component (see section 7.1.3).

**Material**. An object (environment) whose state space is a chain (see section 7.1).

**Policy**. A mapping from states to actions; the formalization of an agent's control structure (see section 5).

**Projection**. A mapping from the state space of one environment to the state space of another (see section 5.2).

**Simple projection**. A mapping between state spaces that maps state space components of one environment to state space components of another (see section 5.2).

**State component**. (For environments whose state spaces are Cartesian products) An element of an environment's state-tuple (see section 5.1).

**Solution**. A policy solves a DCP from an initial state if, when run from that state, it eventually reaches a goal state (see section 5).

**Tool**. (Roughly) A state component that can be brought to ready state without altering any other state components (see section 7.1.3).

**Uniform reducibility**. (Roughly) $E'$ is uniformly reducible to $E$ if it consists of multiple copies of $E$'s objects (see section 7.2.1).

## Glossary of Notation

$\circ$. Function composition operator: $f \circ g(x) = f(g(x))$.

$\pi$. A projection (p. 122).

$\pi^{-1}$. The inverse of $\pi$, *i.e.* the set of states that map to a given state under $\pi$ (p. 131).

$\pi^-$. (For a simple projection $\pi\colon S' \to S$). A generalized inverse. Since $\pi$ only maps certain components of $S'$ to $S$, $\pi^-(s, s')$ is $s'$ with those components replaced by their corresponding components in $s$ (p. 130).

$\mathcal{A}_\pi$. For a simple reduction $\pi$ from an environment $E'$ to $E$, the function mapping an action $a$ from $E$ to the action that implements it in $E'$ (p. 122).

$\mathcal{C}_n$. The chain-environment of $n$ states (p. 127).

$E$. An environment.

$\exists_{E,E'} G$. $G$ a goal of $E$ and $E'$ uniformly reducible to $E$. The existential goal of $G$ in $E'$: the set of all $E'$-states that map to a goal state under some binding (p. 130).

$E_1 \rightleftharpoons E_2$. The serial product. The Cartesian product of $E_1$ and $E_2$ in which actions from the two environments must be taken separately (p. 120).

$E_1 \,\|\, E_2$. The parallel product. The Cartesian product of $E_1$ and $E_2$ in which actions from the two environments must be taken simultaneously (p. 120).

$L_{E,E'}$. ($E'$ an environment uniformly reducible to $E$) The leftmost-ready binding map from $E'$ to $E$ (p. 131).

$p$. A policy.

$p_{E,G}$. The standard policy for single-material environment $E$ and goal $G$ (p. 127).

$\mathcal{S}$. The singleton environment (the environment with exactly one state). Used to represent a self-resetting tool (p. 128).